\RequirePackage{iftex}
\ifPDFTeX
  \pdfoutput=1
\fi
\ifXeTeX
  \PassOptionsToPackage{xetex}{hyperref}
\fi
\ifLuaTeX
  \PassOptionsToPackage{luatex}{hyperref}
\fi
\PassOptionsToPackage{table}{xcolor}

\documentclass[11pt]{article}

\usepackage[final]{acl}

\usepackage{times}
\usepackage{latexsym}
\usepackage[T1]{fontenc}

\usepackage[utf8]{inputenc}

\usepackage{microtype}

\usepackage{inconsolata}
\usepackage{float}

\usepackage{graphicx}
\usepackage{url}
\usepackage{colortbl}        
\usepackage{array}           
\usepackage{booktabs}        
\usepackage{multirow}        
\usepackage{graphicx}  
\usepackage{arydshln}
\usepackage{booktabs}      
\usepackage{array}
\usepackage{tabularx} 
\usepackage{tabularx}
\usepackage{tikz}
\usepackage{pgfplots}
\usepackage{subcaption}
\usepackage{threeparttable}
\usepackage{pgfplots}
\usepackage{listings}
\usepackage{tcolorbox}       

\usepackage{subcaption}
\usepackage{booktabs}
\usepackage{siunitx}
\usepackage{booktabs}    
\usepackage{enumitem}
\usepackage{amsmath} 
\usepackage{fancyvrb}
\usepackage{graphicx}          
\usepackage{amsmath}
\usepackage{amssymb}
\usepackage{bbm}  
\usepackage{booktabs}
\usepackage{siunitx}
\usepackage{multirow}
\usepackage{makecell}
\usepackage{caption}
\usepackage{subcaption}
\usepackage{tikz}
\usepackage{array}       
\usepackage{fancyvrb}    
\usepackage{ragged2e} 
\usepackage{pgfplots}
\usepackage{pgfplotstable}

\title{The Model Agreed, But Didn’t Learn:\\
Diagnosing Surface Compliance in Large Language Models}

\author{
  Xiaojie Gu\textsuperscript{1}\thanks{Equal contribution.},
  Ziying Huang\textsuperscript{1}\footnotemark[1],
  Weicong Hong\textsuperscript{2},
  Jian Xie\textsuperscript{3},
  Renze Lou\textsuperscript{4},
  Kai Zhang\textsuperscript{3} \\
  \textsuperscript{1}Independent Researcher 
  \textsuperscript{2}Cornell Tech \\
  \textsuperscript{3}The Ohio State University 
  \textsuperscript{4}The Pennsylvania State University \\
  \href{mailto:peettherapynoys@gmail.com}{\texttt{peettherapynoys@gmail.com}}
}

\begin{document}
\maketitle


\begin{abstract}

Large Language Models (LLMs) internalize vast world knowledge as parametric memory, yet inevitably inherit the staleness and errors of their source corpora. Consequently, ensuring the reliability and malleability of these internal representations is imperative for trustworthy real-world deployment.
Knowledge editing offers a pivotal paradigm for surgically modifying memory without retraining.
However, while recent editors demonstrate high success rates on standard benchmarks, it remains questionable whether current evaluation frameworks that rely on assessing output under specific prompting conditions can reliably authenticate genuine memory modification.
In this work, we introduce a simple diagnostic framework that subjects models to discriminative self-assessment under in-context learning (ICL) settings that better reflect real-world application environments, specifically designed to scrutinize the subtle behavioral nuances induced by memory modifications.
This probing reveals a pervasive phenomenon of \textit{Surface Compliance}, where editors achieve high benchmark scores by merely mimicking target outputs without structurally overwriting internal beliefs. 
Moreover, we find that recursive modifications accumulate representational residues, triggering cognitive instability and permanently diminishing the reversibility of the model's memory state.
These insights underscore the risks of current editing paradigms and highlight the pivotal role of robust memory modification in building trustworthy, long-term sustainable LLM systems.
Code is available at \url{https://github.com/XiaojieGu/SA-MCQ}.

\end{abstract}

\section{Introduction}
Recent advances in large language models (LLMs)~\cite{llama3,gptj,mistral} have shown that pre-training on massive scale corpora allows models to acquire broad factual and commonsense knowledge implicitly encoded within their parameters~\cite{gpt3_llm_Foundations,safety_alignment,llm_resist}. 
However, this storage mechanism is inherently static, crystallizing the inconsistencies, staleness, and errors present in the source data at the moment of convergence~\cite{negative}. As models are increasingly integrated into dynamic, real-world environments, the capability to selectively modify these internal memory states has become a critical imperative.
Knowledge editing~\cite{memit,rome_counterfact} has emerged as a pivotal paradigm to circumvent this rigidity. 
These techniques aim to surgically intervene in the parameter space, allowing for the precise update of specific internal memory states without retraining or compromising the integrity of unrelated representations.

\begin{figure}[!t]
    \centering
    \includegraphics[width=1\linewidth]{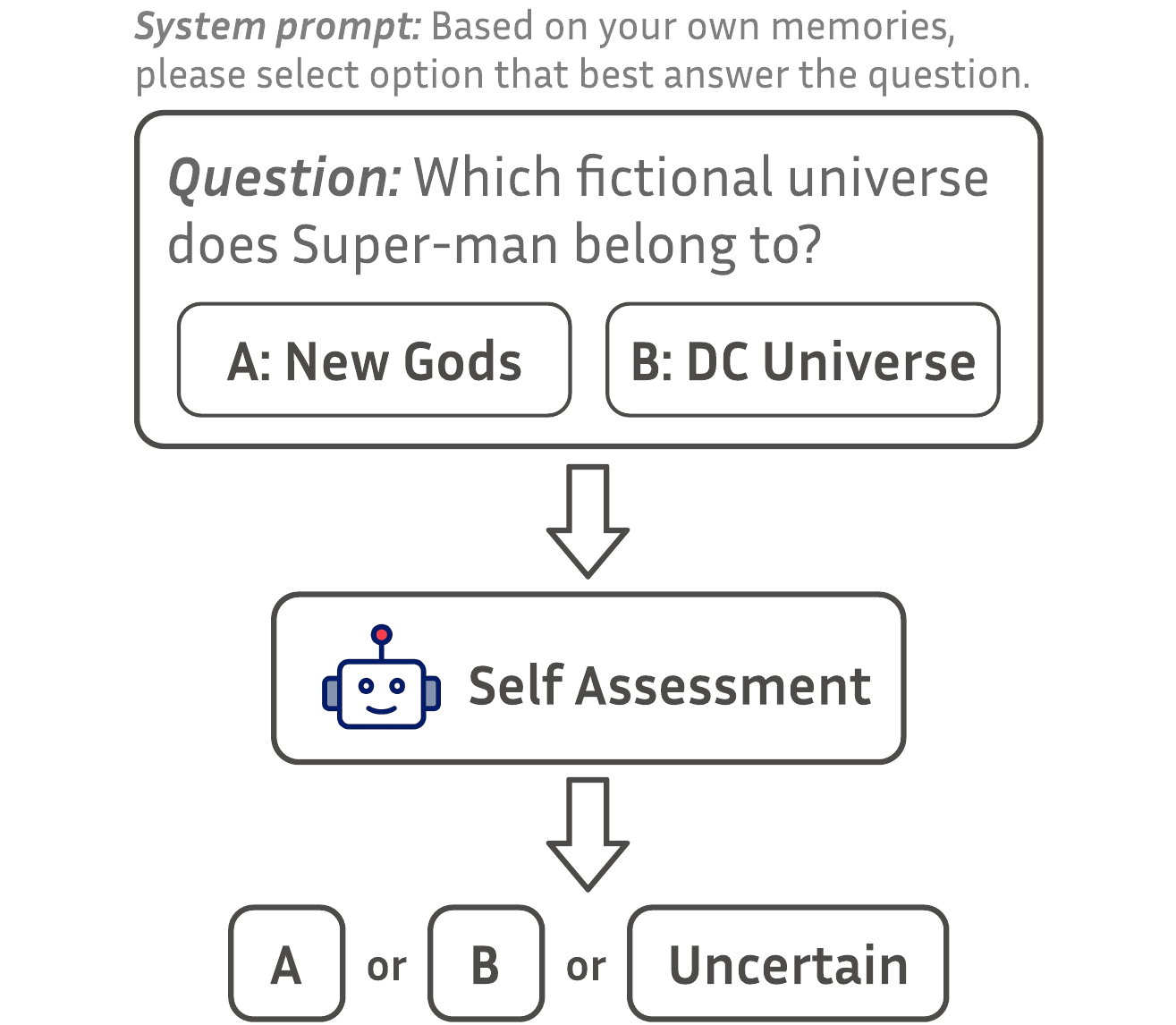}
    \caption{Illustration of the SA-MCQ.}
    \label{SAMCQ}
\end{figure}

\begin{figure}[!t]
    \centering
    \includegraphics[width=1\linewidth]{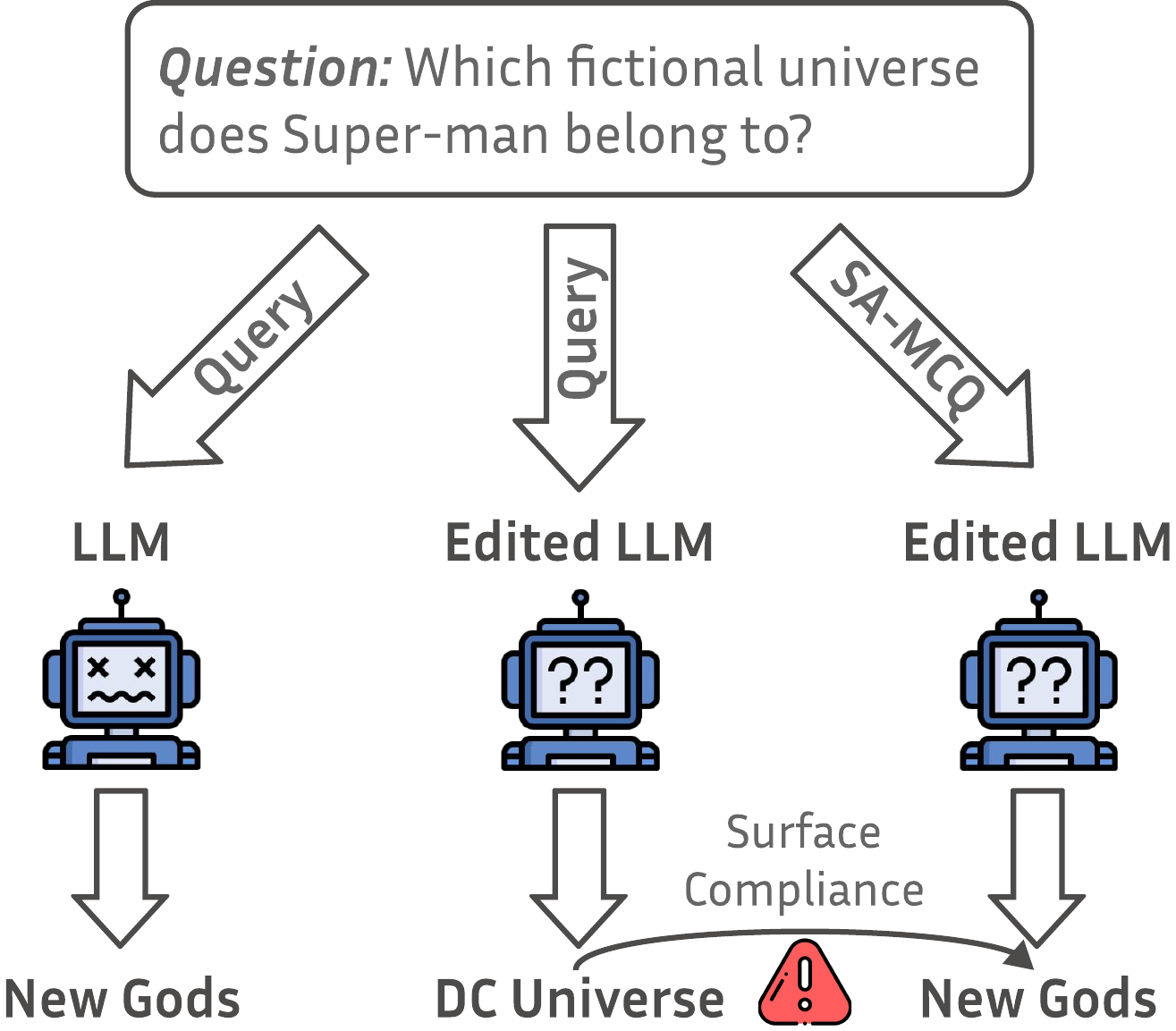}
    \caption{
    Illustration of \textit{Surface Compliance}: Although the edited LLM successfully generates the target golden answer "DC Universe" in traditional evaluation frameworks, it reverts to the parametric answer "New Gods" in the SA-MCQ setting, which probes the genuineness of the memory modification.
    } 
    \label{Surface_Compliance}
\end{figure}

Despite rapid progress, such as achieving million-scale precise editing~\cite{ultraedit},
current evaluation frameworks remain largely confined to free-form generation metrics like Exact Match~\cite{easyedit}. 
These metrics merely assess whether a model can reproduce target tokens under specific prompts, often benefiting from subtle in-context cues that guide the output~\cite{wild}. 
This raises a fundamental question: does surface-level textual agreement truly reflect that the model has learned to reconfigure its internal memory?
To rigorously probe the genuineness of memory modifications, we introduce the Self-Assessment Multiple Choice Question (SA-MCQ) framework (Figure~\ref{SAMCQ}).
By compelling the model to actively adjudicate among competing options, 
SA-MCQ circumvents the rote completion bias~\cite{measure_consistency} inherent in open-ended generation, serving as a discriminative stress test for the modified memory.
Crucially, this diagnostic reveals a critical disconnect we term \textit{Surface Compliance} (illustrated in Figure~\ref{Surface_Compliance}). 
In this state, editors achieve high scores on standard benchmarks yet fail to manifest the corresponding belief change in discriminative assessments. 
This indicates that the model is merely mimicking the target behavior without structurally overwriting its parametric memory.

    


We further extend the SA-MCQ evaluation by designing diverse external evidence, including irrelevant noise and counterfactual scenarios, to emulate the complex in-context dynamics characteristic of real-world deployment. 
This investigation enables us to deeply probe the internal fragility of modified memory, exposing behavioral nuances that emerge only under varying contexts.
Within the field of knowledge editing, these insights underscore the imperative to establish rigorous evaluation frameworks capable of verifying genuine editing efficacy, and subsequent research needs to go beyond simple injection into vanilla models, dedicating increased attention to the efficient and precise iteration of modified memory.
On a broader scale, our findings also provide insights into in-context learning by showing how model behavior can be sharply reshaped by contextual support, conflict, and distraction even when internal memory modification remains incomplete.
Diagnosing such latent inconsistencies is paramount for LLM Trustworthiness and Safety. By distinguishing genuine memory reconfiguration from superficial compliance, our work serves as a vital step towards developing reliable, self-evolving systems resilient to dynamic environments.

Together, we highlight key findings as follows:

\begin{itemize}[leftmargin=10pt]

    \item Surface-level token matching often fails to signify genuine memory reconfiguration, instead masking a fragile state of in-context hypersensitivity. 
    Notably, external counterfactuals can easily suppress the modification, locking the model into a cognitive deadlock.

    \item Recursive memory modification accumulates persistent representational residues that permanently diminish the reversibility of the memory state, triggering cognitive instability and preventing the consolidation of evolving knowledge.

\end{itemize} 

\section{Related Work}

\subsection{Knowledge Editing}
Knowledge editing modifies a model’s stored knowledge to integrate new information without degrading performance, typically following three paradigms~\cite{easyedit}. 
Locate-then-edit methods~\cite{rome_counterfact,memit,alphaedit} identify factual encoding sites via causal tracing or attribution and apply iterative updates to specific layers. While precise for small batches, they suffer from parameter interference during repeated modifications~\cite{prune}.
Memory-augmented approaches~\cite{grace,wise} store new knowledge externally to improve traceability, yet face routing overhead and scalability issues as edits increase. 
Meta-learning editors~\cite{mend,ultraedit,rledit,horse} map edit instructions into weight updates for rapid deployment. 
However, as edits accumulate, they often lead to catastrophic forgetting~\cite{pmo}.

\subsection{Memory in Language Models} 

Large Language Models (LLMs) principally rely on parametric memory, where vast amounts of factual knowledge are implicitly encoded within high-dimensional weight matrices during pre-training ~\cite{language_base1, language_base2}. 
While this distributed representation allows for robust information retrieval, it inherently suffers from rigidity, making models prone to hallucination or obsolescence when world knowledge evolves ~\cite{halsurvey1,halsurvey2}. 
To mitigate this, recent advances in Knowledge Editing have proposed mechanisms to surgically update specific memory slots without retraining, typically by locating and modulating key neuron activations associated with factual associations ~\cite{rome_counterfact, memit}. 
However, consolidating these post-hoc updates remains challenging. Unlike robust pre-trained knowledge, edited memories often exhibit instability, prone to generalization failures ~\cite{pmo} or reversion to priors under interference ~\cite{MQuAKE}.

\section{Preliminary}

\subsection{Editing Paradigm}
The task of knowledge editing concerns the controlled modification of a pre-trained language model so that it adopts new factual associations without retraining from scratch or erasing unrelated capabilities.  
Let $f_\theta : \mathcal{X} \to \mathcal{Y}$ denote a language model with parameters $\theta$.  
An \emph{editing instance} is defined as a pair $(x_e, y_e)$, where $x_e$ denotes a natural language query and $y_e$ is the desired answer. After applying an editing operation, the updated model $f_{\theta'}$ should return $y_e$ when prompted with $x_e$, which reflects \emph{Efficacy}. 
In addition to this direct requirement, the edit is also evaluated on \emph{equivalent instances} $(x_e)$, consisting of paraphrases or semantically similar queries that should likewise yield $y_e$, capturing the dimension of \emph{Generalization}. Finally, the model must retain its original predictions on \emph{unrelated instances}, which are inputs not associated with the edited fact, thereby ensuring \emph{Specificity} and guarding against unintended side effects. 
Examples of three instances and their corresponding metrics \emph{Efficacy (Eff.), Generalization (Gen.), and Specificity (Spe.)} can be found in Appendix \ref{Metric}, where their computation process are detailed.

In this paper, we focus on \emph{sequential editing} (also known as \emph{lifelong editing}), which represents one of the most prominent and practically significant tasks in the field. In this setting, successive modifications are applied cumulatively, with each round building upon all previous edits.
Formally, the model is updated across a sequence of turns: at turn $t$, a collection of edits $\{(x_e^{(t,i)}, y_e^{(t,i)})\}_{i=1}^n$ is applied to the current model $f_{\theta^{(t-1)}}$, resulting in an updated parameterization $f_{\theta^{(t)}}$.

We select three recently proposed and widely used editor, namely AlphaEdit~\cite{alphaedit}, RLEdit~\cite{rledit}, and UltraEdit~\cite{ultraedit}, as representative baselines of current mainstream editing paradigms.
Vanilla represents the original, unedited model.
We do not include memory-augmented methods, 
because these approaches store edited knowledge in external memory components rather than modifying the model’s internal parametric representations, which does not align with our objective.
A detailed description of the editors can be found in Appendix \ref{Editor_details}.

\subsection{Traditional Evaluation Framework}

Traditional evaluation frameworks for knowledge editing mainly fall into three categories: \textit{Exact Match without Teacher Forcing (TF)}, \textit{Exact Match with TF}, and \textit{LLM-as-Judge}. The following provides a detailed description of each.

\noindent\textbf{Exact Match (EM)} measures whether the model’s generated answer exactly matches the reference. It reflects the model’s autoregressive output behavior but is sensitive to minor variations such as synonyms or formatting differences, which can cause semantically correct responses to be marked as incorrect.
As a result, EM may distort a model’s true capabilities.  
This original evaluation setting is referred to as \textit{Exact Match w/o TF}.

\noindent\textbf{Teacher forcing (TF)} is originally a training strategy, yet it has been widely adopted in knowledge editing evaluations~\cite{alphaedit,rledit,ultraedit}. 
In this setup, the model generates tokens by conditioning on gold answer tokens prefixes rather than its own prior outputs.
Empirical studies~\cite{wild} show that TF yields substantially higher accuracy than autoregressive decoding, systematically inflating results and providing an overly optimistic estimate of editing success. 
This setting is often referred to as \textit{Exact Match w/ TF}.

\noindent\textbf{LLM-as-judge} utilizes a stronger language model to grade edited responses against gold targets, accommodating semantic variations that exact matching might miss. 
Despite its flexibility, this framework is computationally expensive and highly sensitive to prompt engineering~\cite{llm_judge_3}.
However, most knowledge editing datasets consist of short-form QA, where editing targets typically comprise fewer than 5 tokens. Despite this, evaluation templates~\cite{wild} often truncate at 512 tokens. Requiring an LLM to evaluate such disproportionate sequences introduces significant bias~\cite{llm_judge_2}.
Furthermore, this framework remains restricted to assessing surface-level text, failing to verify whether the edited knowledge has truly replaced conflicting internal representations.
In our experiments, we follow~\cite{wild} and adopt the instruction-shot template. The specific template is provided in Table~\ref{LLM-as-judge} in Appendix~\ref{Templates}.




\section{Beyond Surface Compliance: Probing Latent Memory Dynamics}

\definecolor{softblue}{RGB}{240, 248, 255} 
\definecolor{highlightblue}{RGB}{176,196,222}
\definecolor{highlightred}{RGB}{255,160,122}

\begin{table*}[!t]
\centering

\setlength{\tabcolsep}{6pt} 
\resizebox{0.9\linewidth}{!}{
\begin{tabular}{lccc ccc ccc ccc}
\toprule
\multirow{2}{*}{Editor} 
& \multicolumn{3}{c}{Exact Match w/ TF} 
& \multicolumn{3}{c}{Exact Match w/o TF} 
& \multicolumn{3}{c}{LLM-as-judge} 
& \multicolumn{3}{c}{Likelihood Margin} \\ 
\cmidrule(l){2-4} \cmidrule(l){5-7} \cmidrule(l){8-10} \cmidrule(l){11-13}
& Eff. & Gen. & Spe. 
& Eff. & Gen. & Spe. 
& Eff. & Gen. & Spe. 
& $\Delta_{\text{edit}}$ & $\Delta_{\text{equiv}}$ & $\Delta_{\text{unrel}}$ \\ 
\midrule
Vanilla 
&45.74   &44.86   &38.21   
&18.10   &16.30   &16.20   
&34.00   &33.90   &44.40   
&-   &-   &-   \\
AlphaEdit 
&96.16   &92.09   &33.03   
&78.00   &67.80   &12.40   
&84.40   &77.80   &42.10   
&14.40   &10.94   &\cellcolor{highlightblue}6.32   \\
RLEdit 
&93.60   &89.38   &49.19   
&72.50   &64.80   &14.20   
&58.90   &58.40   &41.00   
&\cellcolor{highlightred}8.17   &6.65   &4.71   \\
UltraEdit 
&89.88   &83.06   &46.54   
&58.50   &49.20   &16.70   
&56.60   &52.00   &44.60   
&2.95   &0.99   &2.46   \\
\bottomrule
\end{tabular}
}
\caption{Performance of different editors.
Higher Eff., Gen., and Spe. denote indicate better performance. 
For likelihood margins, larger $\Delta_{\text{edit}}$ and $\Delta_{\text{equiv}}$ signify that the model amplifies the golden answer while suppressing the parametric one. In contrast, a smaller $\Delta_{\text{unrel}}$ indicates minimal drift on unrelated memory.}

\label{em_tf_judge}
\end{table*}

In this section, we evaluate whether knowledge editing techniques achieve genuine updates to the model’s underlying parametric memory and examine the subsequent consequences of such changes.

\subsection{Another View of Evaluation}

As illustrated in Figure~\ref{SAMCQ}, traditional evaluation protocols serve as mere surface-level proxies. 
By prioritizing output matching, they conflate shallow alignment with genuine knowledge internalization.
Such metrics fail to reveal whether a edited model can resolve internal conflicts during in-context learning (ICL) in real-world application scenarios, 
ultimately capturing \textbf{\textit{what the model can be nudged to say rather than whether its parametric memory has been genuinely modified}}.


To investigate this further, we introduce \textit{Likelihood Margin} to detect shifts in the edited model's underlying probability distribution. We compute log-likelihood margins across three distinct categories  of instances:

\begin{align}
\Delta_{\text{edit}}(x_e,y_e)
&= \log P_{\theta'}(y_e \mid x_e) \label{eq:edit} \\
&\quad - \log P_{\theta'}(f_{\theta}(x_e) \mid x_e), \nonumber \\[4pt]
\Delta_{\text{equiv}}(x_e',y_e)
&= \log P_{\theta'}(y_e \mid x_e') \label{eq:equiv} \\
&\quad - \log P_{\theta'}(f_{\theta}(x_e) \mid x_e'), \nonumber \\[4pt]
\Delta_{\text{unrel}}(x_u,y_u)
&= \bigl| \log P_{\theta'}(f_{\theta}(x_u) \mid x_u) \label{eq:unrel} \\
&\quad - \log P_{\theta}(f_{\theta}(x_u) \mid x_u) \bigr|. \nonumber 
\end{align}

However, likelihood margins only capture local probability shifts and do not ensure true integration of new knowledge~\cite{turst_memory}. 
Prior studies~\cite{editing_confidence,edingknowledge_cover} demonstrate that varying in-context framing can readily elicit original facts even after editing. 
To address these limitations, we introduce the \textit{Self-Assessment Multiple Choice Question (SA-MCQ)}, a lightweight evaluation protocol (see Table~\ref{efficiency} in Appendix~\ref{Experiments} for efficiency comparison). Unlike generative tasks, this discriminative method serves as a stress test by placing competing knowledge representations, specifically the original parametric memory and the modified target, into direct conflict within the same context. Leveraging this protocol, we formalize the discrepancy between generation and discrimination as \textit{Surface Compliance}: a phenomenon where the edited model correctly generates the golden answer (under the EM w/o TF setting) yet fails to select it in the SA-MCQ setting. This divergence indicates that success in free-form generation often relies on surface-level recall, masking the fact that the underlying parametric memory has not been genuinely modified.
Specifically, this protocol probes whether editing reshapes the model’s internal belief structure beyond probability shifts using two modes. The three-choice mode (\textit{w/ U.}) presents the original parametric answer, the edited target (golden answer), and an uncertain option to expose how the model balances conflicting attractors through indecision. In contrast, the two-choice mode (\textit{w/o U.}) eliminates the uncertainty option to force a commitment, thereby revealing the model's dominant internal preference.
Finally, to mitigate concerns regarding the model’s sensitivity to prompt phrasing and option ordering (positional bias)~\cite{selector}, we perform rigorous sensitivity analyses across various permutations, as detailed in Table~\ref{variants_rlt} in Appendix~\ref{Experiments}.
In SA-MCQ experiments, we use the editing instances pairs.

In addition, we introduce external evidence as controlled stimuli to examine how edited models respond under varying informational contexts. This design enables a deeper investigation into whether editing induces genuine internal reorganization of memory and how the model negotiates, integrates, or resists conflicting signals.
These evidence conditions are constructed as follows:  

\begin{itemize}[leftmargin=10pt]
    \item \textbf{Parametric Evidence (PE)}: 
    We sample questions from the \texttt{{ZsRE}}~\cite{zsre} (Zero-shot Relation Extraction) dataset. In the first step, these questions are queried to the vanilla model to elicit both answers and passages derived from its parametric memory. The resulting answers and passages reflect the model’s original parametric knowledge and approximate the content it would naturally generate without any external assistance.
    We then perform an \textit{Answer Consistency Check}. 
    The extracted passages are fed back to the model as external evidence, after which the model is asked the same question again. If the answer produced in this second round is consistent with the original closed-book answer, the corresponding memory is classified as a firm belief and retained as Parametric Evidence (PE). If the answer is inconsistent, the memory is regarded as unstable and, therefore, discarded.
    
    
    \item \textbf{Golden Evidence (GE)}: 
    We use the same set of questions employed for generating Parametric Evidence (PE) and leverage an external large model to generate passages that support the original annotated gold answers (editing targets) provided with the dataset. These passages, referred to as Golden Evidence (GE), are designed to be factually accurate and tightly aligned with the target concepts.
    

    \item \textbf{Irrelevant Evidence (IE)}: Randomly sample five subject-relation-object triples from UltraEditBench~\cite{ultraedit} and expand them into coherent passages using external large language model. Sentence-BERT~\cite{sentence_bert} is then applied to compute semantic similarity between these passages and other evidence types, and the three passages with the lowest similarity scores are selected as the final Irrelevant Evidence (IE). Although these sentences may appear semantically plausible, they contain no factual connection to the target knowledge and serve as controlled noise to test the model’s robustness against distraction.

    \item \textbf{Counter Evidence (CE)}: 
    We use counterfactual answers from the \texttt{{ZsRE}} dataset and expand them into fluent and coherent passages using an external large language model. These passages, referred to as Counter Evidence (CE), are constructed to explicitly contradict both the Parametric Evidence (PE) and the Golden Evidence (GE), enabling us to examine the model’s behavior when confronted with direct factual conflict.
    
    
\end{itemize}

To mitigate the risk of evidence hallucinations or semantic ambiguity, we incorporate a \textit{Logical Entailment Check} as a prerequisite for our experiments. Leveraging the NLI model \texttt{DeBERTa}~\cite{deberta}, we systematically validate the logical relationship between the generated evidence and their corresponding answers across all four evidence categories. For PE, GE, and CE, we enforce a strict requirement that the evidence explicitly entails the corresponding answer to ensure supportiveness. 
In contrast, for IE, we verify that the content remains logically disconnected from the golden answers to confirm irrelevance. 
To further ensure the reliability of this automated process, we manually evaluate 100 random samples and observe a model accuracy of 98\%, confirming the high fidelity of our data filtering protocol.
Ultimately, we obtain 1K example sets, each of which contains four types of evidence whose corresponding answers are mutually distinct. All experiments reported in the main paper adopt \texttt{LLaMA-3-8B-Instruct}~\cite{llama3} as the backbone model and use \texttt{DeepSeek-V3.2~\cite{deepseekv32}} as the external large language model.
Additional details on the dataset, backbone model selection, and experimental setup are provided in Appendix~\ref{Dataset} and Appendix~\ref{Experiment_details}. 
Example evidence case and the templates used for evidence generation are presented in Table~\ref{case} and Appendix~\ref{Templates}, respectively.



\begin{figure*}[t]
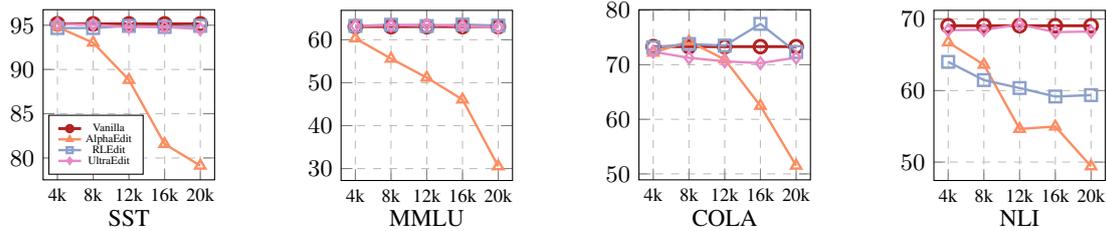

  \centering
  \begin{subfigure}[b]{.235\textwidth}
    \centering
    \input{tables/eval_bench/sst}
  \end{subfigure}\hspace{0.01\textwidth}%
  \begin{subfigure}[b]{.235\textwidth}
    \centering
    \input{tables/eval_bench/MMLU}
  \end{subfigure}\hspace{0.01\textwidth}%
  \begin{subfigure}[b]{.235\textwidth}
    \centering
    \input{tables/eval_bench/COLA}
  \end{subfigure}\hspace{0.01\textwidth}%
  \begin{subfigure}[b]{.235\textwidth}
    \centering
    \input{tables/eval_bench/NLI}
  \end{subfigure}

 \caption{Performance of different edited models as the number of edits increases across various benchmarks.}

  \label{eval_bench}
\end{figure*}

\begin{figure}[h]
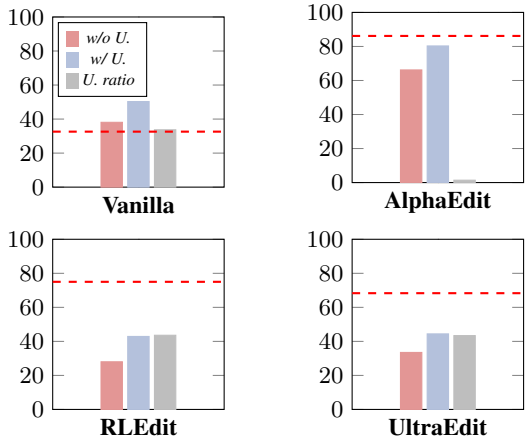

  \centering
  \vspace{-0.5em}

  \begin{subfigure}[t]{0.235\textwidth}
    \centering
    \input{tables/base_main}
  \end{subfigure}\hfill
  \begin{subfigure}[t]{0.235\textwidth}
    \centering
    \input{tables/alpha_main}
  \end{subfigure}\hfill
  \begin{subfigure}[t]{0.235\textwidth}
    \centering
    \input{tables/rl_main}
  \end{subfigure}\hfill
  \begin{subfigure}[t]{0.235\textwidth}
    \centering
    \input{tables/ultra_main}
  \end{subfigure}

  \caption{Ratio of golden answer and uncertain option under SA-MCQ. The red dashed line denotes the mean $Eff.$ obtained under the three traditional evaluation.}
  \label{SA_MCQ_0000_main}
\end{figure}




\subsection{Misalignment between Traditional Metrics and Memory Modification}

We scrutinize the authenticity and stability of memory modification by contrasting experimental outcomes from traditional evaluation frameworks against those of SA-MCQ.

\textbf{Surface-level output compliance and shifts in token probabilities often do not guarantee the genuine modification of pre-trained memory}.
Table \ref{em_tf_judge} implies potential fragility as performance collapses without guiding inputs (performance \textit{EM w/ TF} against \textit{EM w/o TF} and \textit{LLM-as-judge}), the illusion of surface compliance is definitively exposed by contrasting these traditional metrics with the SA-MCQ framework in Figure \ref{SA_MCQ_0000_main}. 
The substantial gap between the average traditional performance (indicated by the red dashed line) and the actual golden selections confirms that surface-level text outputs do not reflect genuine internal memory modifications, leading existing protocols to severely overestimate editing effectiveness.
Extending this scrutiny to the probability level, likelihood analysis reveals that even substantial shifts in token probabilities can be misleading.
A prime example is RLEdit, which induces observable probability changes (high $\Delta_{\text{edit}}$) yet fails to maintain this preference in SA-MCQ. 
This disconnect demonstrates that mere probability shifts are insufficient to guarantee the stable integration of new memory, leaving the model unable to resolve underlying parametric conflicts under discriminative stress.

\textbf{Modifying memory often causes severe collateral damage and can also trigger cognitive instability}.
While AlphaEdit maintains a high golden answer selection rate in Figure \ref{SA_MCQ_0000_main}, this gain is offset by significant interference on unrelated memory (high $\Delta_{\text{unrel}}$ in Table \ref{em_tf_judge}). 
This trade-off is further confirmed by the evaluation on standard benchmarks (MMLU~\cite{mmlu}, GLUE~\cite{glue}, NLI~\cite{nli}, etc.) in Figure \ref{eval_bench}, where AlphaEdit causes a notable, progressive degradation in performance as the number of edits increases.
Furthermore, RLEdit and UltraEdit exhibit golden selection rates even lower than the vanilla model, and yielding a marked surge in the proportion of uncertain choices. 
This indicates that these methods damage the model’s internal stability without successfully integrating the target memory.
For the complete set of experimental results, please refer to Figures~\ref{SA_MCQ_0000_llama}, \ref{SA_MCQ_0000_qwen}, and \ref{eval_bench2} in Appendix~\ref{Experiments}.

\begin{figure*}[t]
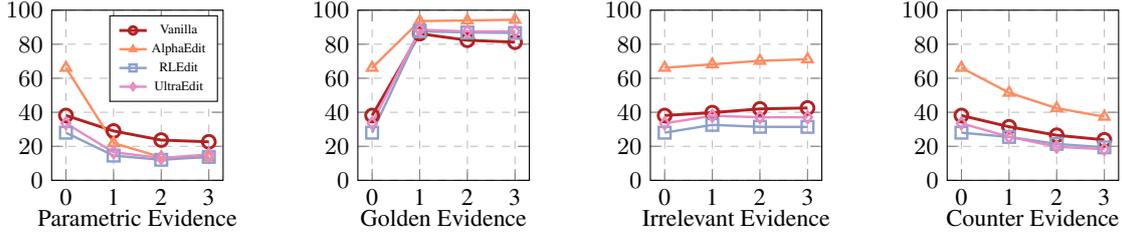

  \centering
  \begin{subfigure}[b]{.235\textwidth}
    \centering
    \input{tables/pe_num}
  \end{subfigure}\hspace{0.01\textwidth}%
  \begin{subfigure}[b]{.235\textwidth}
    \centering
    \input{tables/ge_num}
  \end{subfigure}\hspace{0.01\textwidth}%
  \begin{subfigure}[b]{.235\textwidth}
    \centering
    \input{tables/ie_num}
  \end{subfigure}
  \begin{subfigure}[b]{.235\textwidth}
    \centering
    \input{tables/ce_num}
  \end{subfigure}\hspace{0.01\textwidth}%

\caption{Ratio of golden answer choice under the SA-MCQ protocol in the $w/o\ U.$ setting. Results are evaluated on \textit{Surface Compliance} instances, as the amount of external evidence increases across different editors.}
\label{evidence_num}
\end{figure*}

\subsection{Effect of External Evidence}

We introduce varying evidence to investigate the in-context sensitivity of memory-modified model.

\textbf{Memory modification renders the model sensitive to external contexts aligning with either the original pre-trained memory or the target modification}, irrespective of whether the modification itself is genuinely successful.
As shown in Figure~\ref{evidence_num}, under the Parametric Evidence setting, the golden answer selection ratio decreases across all editors (indicating a regression to the parametric answer), with the magnitude of this shift surpassing that of the vanilla model. Conversely, under the Golden Evidence setting, all editors exhibit a sharper increase in golden answer selection compared to the vanilla baseline. Notably, even editors like RLEdit and UltraEdit, which previous analyses identified as having failed to genuinely integrate the target memory, display this heightened sensitivity. 
This also suggests that current editing mechanisms primarily function by disrupting the inertia of the original parametric weights rather than precisely overwriting them.

\textbf{Successful memory modification fortifies the model's resistance to irrelevant noise}, whereas ineffective modification renders the model susceptible to distraction.
Observations in the Irrelevant Evidence setting substantiate this distinction: AlphaEdit maintains a golden selection rate significantly superior to the vanilla model, demonstrating its capability to disregard noise. In contrast, RLEdit and UltraEdit perform slightly below the vanilla baseline, indicating a lack of such robustness. Furthermore, the accumulation of external context exacerbates this divergence. As the number of evidence pieces increases, the performance gap between the three editors and the vanilla model widens.

\textbf{External counterfactual context can easily suppress the memory modification, forcing models into a state of cognitive deadlock}.
Under the Counter Evidence setting, where the context contradicts both the original parametric memory and the target, the preference for the golden answer declines significantly. This behavioral pattern gradually converges toward the pre-trained vanilla baseline as the volume of evidence increases. Crucially, this suppression of the target is accompanied by a substantial rise in the selection of the uncertain option, as shown in Figure \ref{MCQ_evidence1} in the Appendix. This trend confirms that rather than robustly rejecting the counterfactuals, the models succumb to the heightened dissonance between internal weights and external cues. 
Consequently, the external context overwhelms the internal parametric structure, causing the model to abandon the target knowledge in favor of indecision.

\begin{figure*}[t]
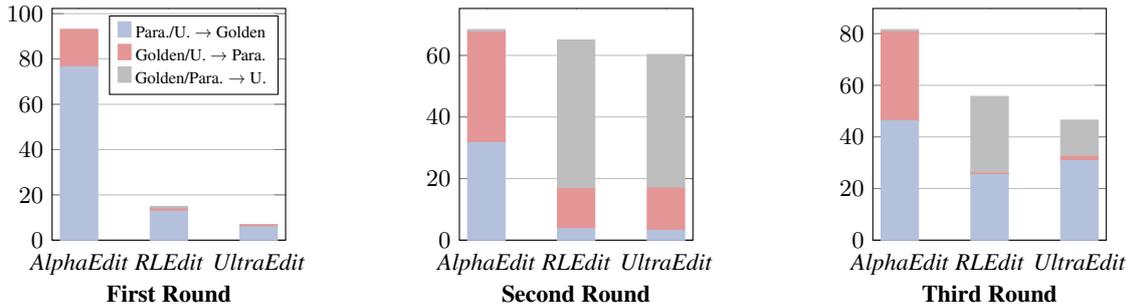

  \centering
  \vspace{-0.5em}

  \begin{subfigure}[t]{0.32\textwidth}
    \centering
    \input{tables/cf_1}
  \end{subfigure}\hfill
  \begin{subfigure}[t]{0.32\textwidth}
    \centering
    \input{tables/cf2}
  \end{subfigure}\hfill
  \begin{subfigure}[t]{0.32\textwidth}
    \centering
    \input{tables/cf3}
  \end{subfigure}

  \caption{Results after three editing rounds. $Para./U.\ \to \ Golden$ denotes the ratio of transitions from parametric or uncertain option to the golden answer relative to the previous round; other legend items follow the same logic.
  The conversion ratios in the \textit{First Round} are evaluated relative to the vanilla model.}
  \label{cf_all}
\end{figure*}

\section{Modifying the Modified: Probing the Plasticity of Memory}


The constant evolution of knowledge necessitates continuous updates, raising the question of whether modified memory retains the receptivity of the vanilla model. We investigate whether re-modification introduces unique structural conflicts compared to the initial process.

\subsection{Multi-Round Editing Design}

To examine this phenomenon, we conduct a controlled three-round editing experiment that alternates between factual and counterfactual modifications. 
This design allows us to trace how successive interventions reshape the model’s internal memory and to assess whether existing editing methods can preserve both consistency and adaptability when repeatedly modifying the edited knowledge.
The three-round editing process is designed as follows.

\begin{itemize}[leftmargin=10pt]

   \item \textbf{First Round}: Starting from the vanilla model, 1K facts are injected to override parametric memory with golden answer.

   \item \textbf{Second Round}:
   Building on the edited model from the First Round, models are further edited on the corresponding 1K counterfactual answers, which are deliberately designed to conflict with the previously injected facts.
   In this round, the counterfactual answers serve as the new editing targets (thus becoming the new golden answers), while the original factual instances from the First Round now act as counter evidence against them.

    \item \textbf{Third Round}: Building on the edited model from the Second Round, the original factual instances (the same in the First Round) are reintroduced to override counterfactual knowledge.

\end{itemize}

\subsection{Re-modification Induces Accumulative Conflict}

Figure \ref{cf_all} reveals that the first round of editing with AlphaEdit effectively modifies the model’s memory, transcending simple output adjustments. Conversely, RLEdit and UltraEdit induce only sparse conversions toward the target memory. 
This empirical evidence reinforces our earlier conclusion regarding the capacity of these editors in achieving genuine memory modification.

\definecolor{softblue}{RGB}{240, 248, 255} 

\begin{table}[h]
\centering

\label{cf_evidence}

\setlength{\tabcolsep}{6pt}

\resizebox{1.0\linewidth}{!}{
\begin{tabular}{l *{5}{c}}
\toprule
Editor
& No evidence
& PE
& GE
& IE
& CE \\
\midrule
\rowcolor[gray]{0.85}
Vanilla      & 33.7 & 14.0 & 88.5 & 17.6 & 31.5 \\
\textit{First Round} \\
\cdashline{1-6}[4pt/4pt]
AlphaEdit & 79.4 &  5.9 & 93.5 & 69.4 & 53.6 \\
RLEdit    & 41.9 & 15.9 & 89.1 & 21.9 & 36.6 \\
UltraEdit & 37.3 & 14.5 & 88.9 & 20.0 & 30.1 \\
\midrule
\textit{Second Round} \\
\cdashline{1-6}[4pt/4pt]
AlphaEdit & 56.6 &  7.5 & 93.8 & 28.0 & 47.6 \\
RLEdit    &  8.4 &  4.0 & 87.4 &  4.5 & 18.1 \\
UltraEdit &  8.0 &  3.5 & 85.8 &  4.0 & 15.1 \\
\midrule
\textit{Third Round} \\
\cdashline{1-6}[4pt/4pt]
AlphaEdit & 56.8 & 13.5 & 96.0 & 64.6 & 60.8 \\
RLEdit    & 27.8 & 17.0 & 89.7 & 16.2 & 27.0 \\
UltraEdit & 34.2 & 14.6 & 88.8 & 17.2 & 30.4 \\
\bottomrule
\end{tabular}
}
\caption{Ratio of golden answer choice across three rounds, evaluated under different evidence scenarios.}

\label{cf_evidence}
\end{table}

\textbf{Artificially implanted memory is significantly more brittle than the model’s pre-trained parameters.}
When counterfactual targets are introduced in the second round, memory stability deteriorates sharply. The surge in transitions reverting to the original parametric answer, coupled with a substantial drop in golden target adherence, reveals that ineffective re-modification compromises the integrity of the previously modified memory. This leaves the model in a metastable state that resists the full consolidation of the target, a fragility further corroborated by the decline in golden preference under the no-evidence condition in Table \ref{cf_evidence}. Furthermore, while reintroducing the target in the third round yields partial recovery, it fails to achieve full stabilization. The persistence of significant uncertainty indicates that the conflicting updates have structurally disrupted the memory, preventing it from settling back into a coherent deterministic state.


\textbf{Conflicts between memory leave persistent representational residues that diminish the reversibility of memory states}.
As shown in Table \ref{cf_evidence} under the No Evidence setting, the golden selection ratios for all editors in the third round fail to recover to the peak fidelity observed in the first round, despite the re-application of the target knowledge. 
Complementing this observation, results under the Irrelevant Evidence setting reveal a parallel degradation in robustness. As established in last section, effective modification typically fortifies resistance to irrelevant external noise; however, even AlphaEdit which demonstrates high modification success exhibits third-round performance that falls distinctly below its first-round levels. 
These cumulative deficits indicate that recursive modification of previously updated memory leaves persistent residual conflicts that obstruct the full restoration of the target equilibrium.
Furthermore, these findings underscore a critical gap in the knowledge editing field. While the prevailing paradigm predominantly targets the injection of new information into vanilla models, it largely overlooks the complexities of  edited knowledge. Consequently, there is an urgent need to extend this scope toward re-editing, specifically developing strategies to mitigate the cumulative negative impacts and structural instability arising from recursive updates.

\section{Conclusion}

Our study reveals that widely used open-ended generation metrics capture \textit{Surface Compliance} rather than genuine parametric reconfiguration. 
We show that without structurally overwriting internal beliefs, edited models remain fragile, accumulating destabilizing residues from recursive updates. 
Addressing these latent inconsistencies is essential for Trustworthiness, ensuring models truly "learn" to adapt in dynamic environments rather than merely "agreeing" with target updates.

\section{Limitations}
Due to computational resource constraints and experimental setup, we are unable to extend the experiments to additional datasets or models with larger parameter scales. A detailed explanation is provided in Appendix~\ref{Dataset}.
\bibliography{custom}

\clearpage

\appendix

\section{Appendix}

\label{appendix}


\subsection{Editing examples and Metric}\label{Metric}

The following are editing examples along with their corresponding metrics.

\begin{itemize}
    \item \textbf{Editing query}: Which fictional universe does Super-man belong to? 
    \item \textbf{Equivalent query}: In which fictional universe does Super-man exist?    
    \item \textbf{Unrelated query}: Who holds the most home runs in MLB history?

\end{itemize}

\textit{Efficacy (Eff.)} measures whether the edit has been faithfully integrated. 
It requires the updated model $f_{\theta'}$ to output the target label $y_e$ when given the edited query $x_e$:  
\begin{equation}
\begin{aligned}
&\mathrm{Efficacy}=  \\
&\mathbb{E}\left[\mathbf{1}\!\left(y^e = \arg\max_{y'} \, \mathbb{P}_{f_{\theta'}}(y' \mid x_e)\right)\right]
\end{aligned}
\end{equation}


\textit{Generalization (Gen.)} examines whether the edit generalizes to equivalent queries. 
For paraphrases $x_e' $, the model should likewise return $y_e$:  
\begin{equation}
\begin{aligned}
&\mathrm{Generalization} = \\
&\mathbb{E}_{x_e' \in \mathcal{E}(x_e)}\left[\mathbf{1}\!\left(y^e = \arg\max_{y'} \, \mathbb{P}_{f_{\theta'}}(y' \mid x_e')\right)\right]
\end{aligned}
\end{equation}

\textit{Specificity (Spe.)} verifies that unrelated knowledge is preserved after editing.. 
For each unrelated input $x_u $, the updated model should retain its original prediction $y_u$:  
\begin{equation}
\begin{aligned}
&\mathrm{Specificity} =\\
&\mathbb{E}_{x_u \in \mathcal{U}(x_e)}\left[\mathbf{1}\!\left(y^u = \arg\max_{y'} \, \mathbb{P}_{f_{\theta'}}(y' \mid x_u)\right)\right].
\end{aligned}
\end{equation}

\subsection{Dataset \& Backbone.}\label{Dataset}
We conduct our experiments on the widely used benchmark for knowledge editing, the \texttt{{ZsRE}}~\cite{zsre} (Zero-shot Relation Extraction) dataset. ZsRE is derived from the original question–answer pairs in the Natural Questions corpus, where each instance is rewritten into a relational query paired with its corresponding factual answer.
In addition to the edited instances, equivalent instances, and unrelated instances, each example in ZsRE is also annotated with a counterfactual answer that provides a plausible yet incorrect alternative entity.
This enables controlled evaluation of both factual recall and conflict resolution, making ZsRE a standard benchmark for testing whether models can correctly retrieve or update entity–relation knowledge after editing.
We exclude \texttt{CounterFact}~\cite{rome_counterfact} because the \texttt{ZsRE} already contains counterfactual annotations. We also exclude \texttt{MQuAKE}~\cite{MQuAKE}, \texttt{EVOKE}~\cite{evoke}, and \texttt{AKEW}~\cite{akew} due to their limited dataset scale.

For the backbone models, we adopt \texttt{LLaMA-3-8B-Instruct}~\cite{llama3} and \texttt{Qwen2.5-7B-Instruct}~\cite{qwen}, both instruction-tuned large language models that provide a strong foundation for evaluating knowledge editing methods.
We exclude the GPT family of models (e.g., \texttt{GPT-J}~\cite{gptj}) because they do not exhibit sufficiently strong instruction-shot capabilities in our setting. 
We also exclude \texttt{Mistral}~\cite{mistral}, as AlphaEdit does not provide corresponding hyperparameter configurations for this model and yields zero performance on this architecture.
Due to computational constraints and the high VRAM requirements of methods like RLEdit (e.g., 79.86 GB for an 8B model), we follow most prior works and limit our experiments to models with no more than 8B parameters.

\definecolor{subj}{HTML}{bfc9df}
\definecolor{parametric_memory}{HTML}{f5ce93}
\definecolor{subjPrime}{HTML}{c2e3bf}
\definecolor{objPrime}{HTML}{bce1e1}
\newcommand{\tableColor}[2][m_ans]{%
  \colorbox{#1}{%
    \parbox[t]{\linewidth}{#2\strut}%
  }%
}

\newcommand{\levelone}{Level-1}
\newcommand{\leveltwo}{Level-2}
\newcommand{\levelthree}{Level-3}

\newcommand{\hlbox}[2]{%
  {\setlength{\fboxsep}{0pt}\colorbox{#1}{#2\strut}}%
}

\newcommand{\tstrut}{\rule{0pt}{1.1em}}

\setlength{\tabcolsep}{4pt}
\renewcommand{\arraystretch}{1.12}
\captionsetup{font=small}

\newcolumntype{Y}{>{\noindent\justifying\arraybackslash}X}

\begin{table*}[h]
  \centering
\label{evidence_case}

\begin{tabularx}{\textwidth}{YYYY}
    \toprule
    \multicolumn{4}{l}{
  \makecell[l]{
    \textbf{Question:} What is the name of the person who discovered 14 Irene? \quad
    \textbf{Parametric Answer:} Aubert \quad \\
    \textbf{Golden Answer:} John Russell Hind  \quad 
    \textbf{Counter Answer}: Karl Ludwig Harding
  }
} \\
    \midrule
    \addlinespace[2pt]

    \tableColor[subj] {\tstrut\textbf{Parametric Evidence:} 
    As I think back to my college days, I recall studying the properties of radioactive isotopes and their applications in medicine and industry. 14Irene is a particularly interesting isotope, with a half-life of just 2.1 million years. I remember Aubert's discovery being a major breakthrough in the field...} 
    &
    \tableColor[subj] {\tstrut\textbf{Golden Evidence:} John Russell Hind, a British astronomer, discovered the asteroid 14 Irene on May 19, 1851. Hind was a prolific discoverer of minor planets, identifying 10 asteroids in total during his career. His work was conducted primarily from the private observatory of George Bishop in London, where he served...}
    &
    \tableColor[subj] {\tstrut\textbf{Irrelevant Evidence:} Glenternie House is a historic country estate located in the Scottish Borders. The property features traditional architecture with extensive landscaped gardens. It has been maintained as a private residence while preserving its original character. The surrounding area offers scenic views...} 
    &
    \tableColor[subj] {\tstrut\textbf{Counter Evidence:} The asteroid 14 Irene was discovered by German astronomer Karl Ludwig Harding in 1851. Harding, known for his work at the Göttingen Observatory, identified the celestial body during a systematic survey of the asteroid belt, adding to the growing catalog of minor planets in the mid-19th century...}


    \\
    \midrule
      \end{tabularx}
\caption{Examples of evidence in the SA-MCQ protocol.}
\label{case}
\end{table*}

\subsection{Editor details}\label{Editor_details}

\noindent\textbf{AlphaEdit}  performs knowledge editing through null-space constrained updates, projecting parameter changes onto directions orthogonal to preserved knowledge. This prevents representation drift and maintains stability across sequential edits while efficiently integrating into existing locate-then-edit pipelines.

\noindent\textbf{RLEdit} formulates model editing as a reinforcement learning problem, treating each parameter update as an action guided by editing rewards. This enables adaptive updates that preserve prior edits and sustain stability over long editing sequences.

\noindent\textbf{UltraEdit} achieves training-, subject-, and memory-free editing by decoupling the editing objective from model weights and using lightweight controller modules for context-aware updates. It generalizes across architectures while mitigating interference from repeated edits.

\subsection{Experiment details}\label{Experiment_details}
All experiments are conducted on a single NVIDIA A800 GPU.

\noindent The editable modules are configured according to the original settings of these methods:

\noindent\textbf{AlphaEdit:} \texttt{[4--8].mlp.down\_proj}

\noindent\textbf{RLEdit:} \texttt{[11--15].mlp.gate\_proj} and 
\texttt{[18--24].mlp.up\_proj}.

\noindent\textbf{UltraEdit:} \texttt{[11--15].mlp.gate\_proj} and 
\texttt{[18--24].mlp.up\_proj}.







\subsection{Experiment Result}\label{Experiments}
We define \textit{Surface Compliance (Sur. Com.)} as the phenomenon where a model generates the golden answer under the EM w/o TF setting but fails to select the corresponding golden option in SA-MCQ. Conversely, \textit{Surface Failure (Sur. Fail.)} refers to the case where the model fails to generate the golden answer in EM w/o TF but successfully identifies the golden option in SA-MCQ.
Experiments that do not explicitly distinguish between \textit{Sur. Com.} and \textit{Sur. Fail.} are conducted exclusively on \textit{Sur. Fail.} instances.

\begin{table}[h]
\centering
\small 
\begin{tabular}{lr}
\toprule
\textbf{Evaluation Framework} & \textbf{Time (min)} \\
\midrule
EM w/o TF                     & 1.82               \\
EM w/ TF                      & 0.05               \\
LLM-as-judge*               & 651.33             \\
Likelihood                 & 1.65               \\
SA-MCQ                     & 0.88               \\
\bottomrule
\end{tabular}
\caption{
Evaluation time comparison across different frameworks.
“*” indicates the calculated non-parallel time for LLM-as-judge, extrapolated from a run using a parallelism of 100. 
Remaining methods also report non-parallel processing times.
}
\label{efficiency}
\end{table}

\begin{table}[h]
    \centering

\resizebox{\columnwidth}{!}{%
    \begin{tabular}{lccccc} 
        \toprule
        & \multicolumn{2}{c}{w/o Uncertain} & \multicolumn{2}{c}{w/ Uncertain} \\ 
        \cmidrule(lr){2-3} \cmidrule(lr){4-5}
        & Golden First & Parametric First & Golden First & Parametric First \\ 
        \midrule
        Variant & $40.7 \pm 2.29$ & $38.4 \pm 3.04$ & $50.0 \pm 2.66$ & $51.3 \pm 1.23$ \\ 
       
        \bottomrule
    \end{tabular}%
    }
\caption{Mean golden selection ratios across three template variants of SA-MCQ on \texttt{LLaMA-3-8B-Instruct} vanilla model. ``Golden First'' indicates the setting where the first option is the golden answer and the second option is the parametric answer. Values following $\pm$ denote the standard deviation.}
\label{variants_rlt}
\end{table}

\begin{figure}[h]
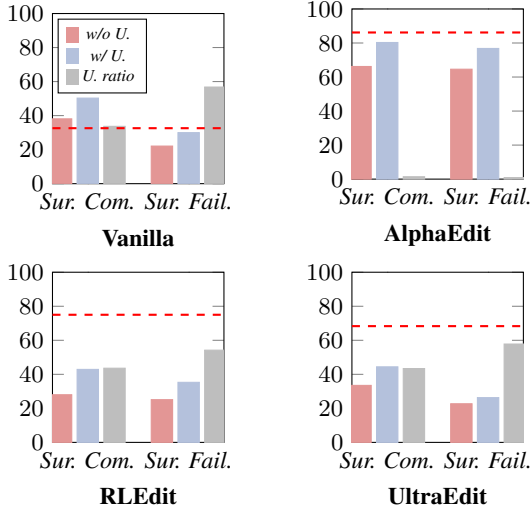

  \centering
  \vspace{-0.5em}

  \begin{subfigure}[t]{0.235\textwidth}
    \centering
    \input{tables/base_suc000}
  \end{subfigure}\hfill
  \begin{subfigure}[t]{0.235\textwidth}
    \centering
    \input{tables/alpha_suc000}
  \end{subfigure}\hfill
  \begin{subfigure}[t]{0.235\textwidth}
    \centering
    \input{tables/rl_suc000}
  \end{subfigure}\hfill
  \begin{subfigure}[t]{0.235\textwidth}
    \centering
    \input{tables/ultra_suc000}
  \end{subfigure}

  \caption{Ratio of golden answer and uncertain option without external evidence. The red dashed line denotes the mean $Eff.$ obtained under the three traditional evaluation.}
  \label{SA_MCQ_0000_llama}
\end{figure}

\begin{figure}[h]
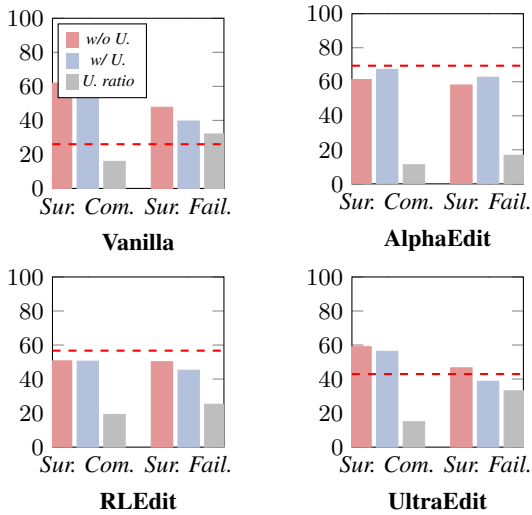

  \centering
  \vspace{-0.5em}

  \begin{subfigure}[t]{0.235\textwidth}
    \centering
    \input{tables/base_suc000_qwen}
  \end{subfigure}\hfill
  \begin{subfigure}[t]{0.235\textwidth}
    \centering
    \input{tables/alpha_suc000_qwen}
  \end{subfigure}\hfill
  \begin{subfigure}[t]{0.235\textwidth}
    \centering
    \input{tables/rl_suc000_qwen}
  \end{subfigure}\hfill
  \begin{subfigure}[t]{0.235\textwidth}
    \centering
    \input{tables/ultra_suc000_qwen}
  \end{subfigure}

  \caption{Ratio of golden answer and uncertain option without external evidence on \texttt{Qwen2.5-7B-Instruct}.
  }
  \label{SA_MCQ_0000_qwen}
\end{figure}

\definecolor{softblue}{RGB}{240, 248, 255} 

\begin{table*}[!t]
\centering

\setlength{\tabcolsep}{6pt} 
\resizebox{0.9\linewidth}{!}{
\begin{tabular}{lccc ccc ccc ccc}
\toprule
\multirow{2}{*}{Editor} 
& \multicolumn{3}{c}{Exact Match w/ TF} 
& \multicolumn{3}{c}{Exact Match w/o TF} 
& \multicolumn{3}{c}{LLM-as-judge} 
& \multicolumn{3}{c}{Likelihood Margin} \\ 
\cmidrule(l){2-4} \cmidrule(l){5-7} \cmidrule(l){8-10} \cmidrule(l){11-13}
& Eff. & Gen. & Spe. 
& Eff. & Gen. & Spe. 
& Eff. & Gen. & Spe. 
& $\Delta_{\text{edit}}$ & $\Delta_{\text{equiv}}$ & $\Delta_{\text{unrel}}$ \\ 
\midrule
Vanilla   
&41.92 &40.32 &38.56 
&13.15 &10.04 &11.90 
&22.89 &21.78 &34.50 
&-    &-    &-     \\
AlphaEdit 
&90.63 &95.07 &43.60 
&63.98 &55.59 &6.63  
&53.73 &49.79 &23.81 
&3.41 &4.32 &14.48 \\
RLEdit    
&86.03 &81.02 &46.11 
&42.86 &37.16 &11.49 
&41.22 &38.11 &31.44 
&3.39 &2.95 &4.74  \\
UltraEdit 
&74.20 &65.13 &39.60 
&23.29 &20.29 &12.84 
&31.33 &27.11 &35.22 
&-7.28 &-7.69 &11.40 \\
\bottomrule
\end{tabular}
}
\caption{Performance of different editors on \texttt{Qwen2.5-7B-Instruct}.}
\label{em_tf_judge_qwen}
\end{table*}

\begin{figure*}[t]
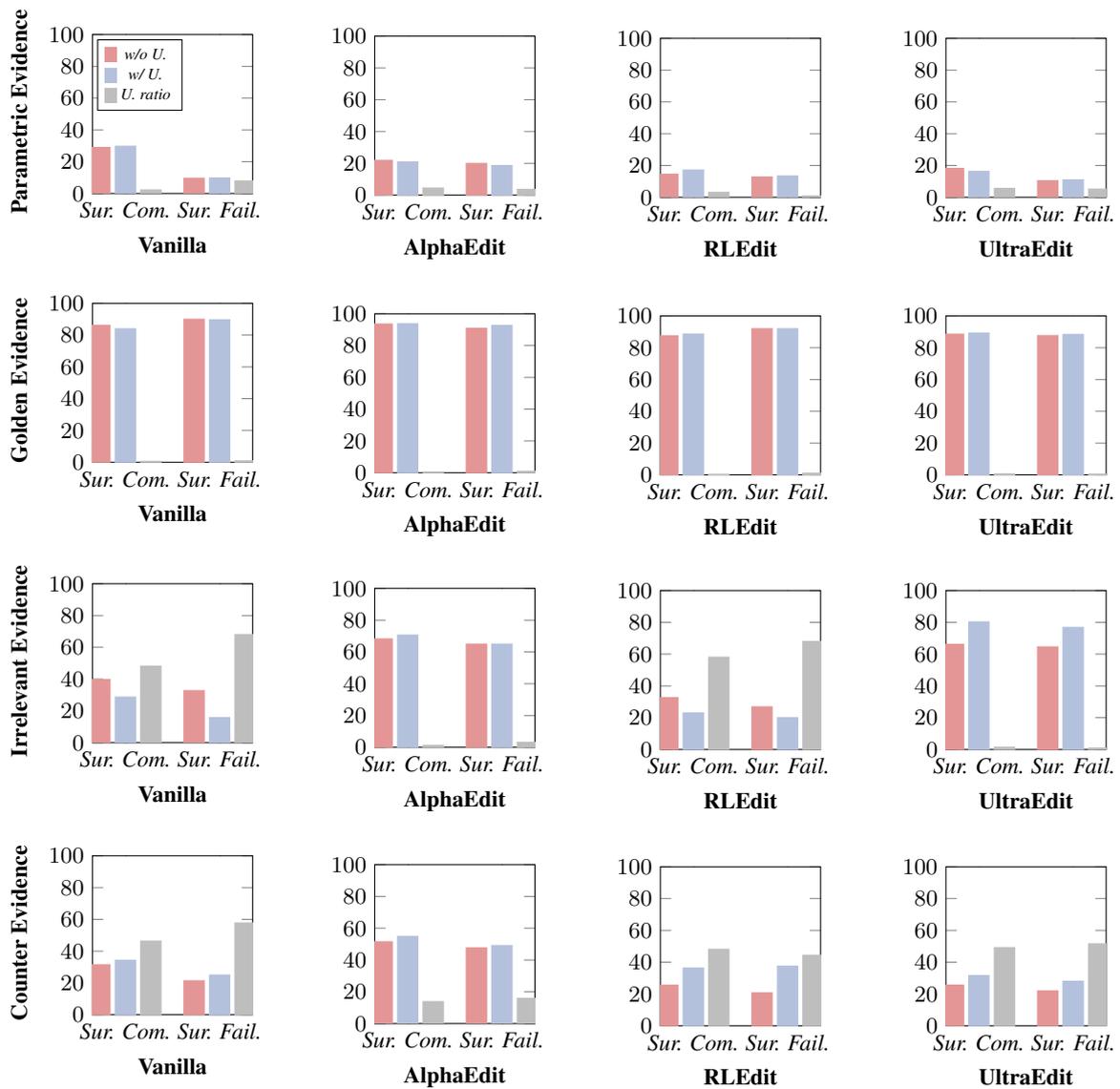

    \centering
     \vspace{-1em}

    \begin{minipage}{.23\textwidth}
        \centering
        \input{tables/base100}     
        \label{fig:ent_vs_gen_chatgpt}
    \end{minipage}\hspace{0.04\textwidth}
    \begin{minipage}{.23\textwidth}
        \input{tables/alpha100}
        \label{fig:ent_vs_gen_gpt4}
    \end{minipage}\hspace{0.01\textwidth}
    \begin{minipage}{.23\textwidth}
        \input{tables/rl100}
        \label{fig:ent_vs_gen_gpt4}
    \end{minipage}\hspace{0.01\textwidth}
    \begin{minipage}{.23\textwidth}
        \input{tables/ultra100}
        \label{fig:ent_vs_gen_gpt4}
    \end{minipage}\hspace{0.01\textwidth}
    \begin{minipage}{.23\textwidth}
        \centering
        \input{tables/base010}     
        \label{fig:ent_vs_gen_chatgpt}
   \end{minipage}\hspace{0.04\textwidth}
    \begin{minipage}{.23\textwidth}
        \input{tables/alpha010}
        \label{fig:ent_vs_gen_gpt4}
    \end{minipage}\hspace{0.01\textwidth}
    \begin{minipage}{.23\textwidth}
        \input{tables/rl010}
        \label{fig:ent_vs_gen_gpt4}
    \end{minipage}\hspace{0.01\textwidth}
    \begin{minipage}{.23\textwidth}
        \input{tables/ultra010}
        \label{fig:ent_vs_gen_gpt4}
    \end{minipage}\hspace{0.01\textwidth}
    \begin{minipage}{.23\textwidth}
        \centering
        \input{tables/base001}     
        \label{fig:ent_vs_gen_chatgpt}
    \end{minipage}\hspace{0.04\textwidth}
    \begin{minipage}{.23\textwidth}
        \input{tables/alpha001}
        \label{fig:ent_vs_gen_gpt4}
    \end{minipage}\hspace{0.01\textwidth}
    \begin{minipage}{.23\textwidth}
        \input{tables/rl001}
        \label{fig:ent_vs_gen_gpt4}
    \end{minipage}\hspace{0.01\textwidth}
    \begin{minipage}{.23\textwidth}
        \input{tables/ultra001}
        \label{fig:ent_vs_gen_gpt4}
    \end{minipage}\hspace{0.01\textwidth}
    \begin{minipage}{.23\textwidth}
        \centering
        \input{tables/base0001}     
        \label{fig:ent_vs_gen_chatgpt}
    \end{minipage}\hspace{0.04\textwidth}
    \begin{minipage}{.23\textwidth}
        \input{tables/alpha0001}
        \label{fig:ent_vs_gen_gpt4}
    \end{minipage}\hspace{0.01\textwidth}
    \begin{minipage}{.23\textwidth}
        \input{tables/rl0001}
        \label{fig:ent_vs_gen_gpt4}
    \end{minipage}\hspace{0.01\textwidth}
    \begin{minipage}{.23\textwidth}
        \input{tables/ultra0001}
        \label{fig:ent_vs_gen_gpt4}
    \end{minipage}\hspace{0.01\textwidth}
    
    \caption{Results of edited models under the SA-MCQ protocol with single evidence. 
}
    \label{MCQ_evidence1}
\end{figure*}

\begin{figure*}[t]
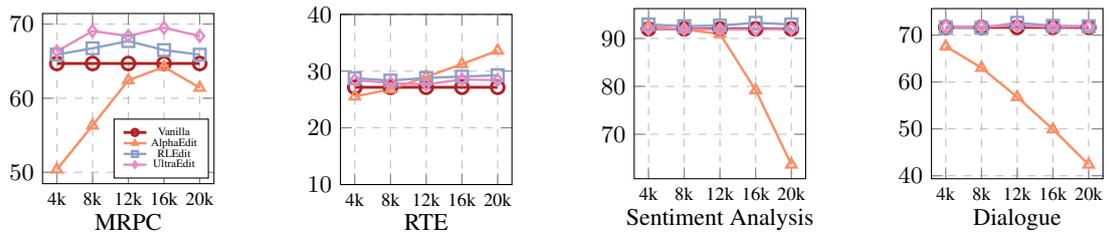

  \centering
  \begin{subfigure}[b]{.235\textwidth}
    \centering
    \input{tables/eval_bench/MRPC}
  \end{subfigure}\hspace{0.01\textwidth}%
  \begin{subfigure}[b]{.235\textwidth}
    \centering
    \input{tables/eval_bench/RTE}
  \end{subfigure}\hspace{0.01\textwidth}%
  \begin{subfigure}[b]{.235\textwidth}
    \centering
    \input{tables/eval_bench/Sentiment_analysis}
  \end{subfigure}
  \begin{subfigure}[b]{.235\textwidth}
    \centering
    \input{tables/eval_bench/Dialogue}
  \end{subfigure}\hspace{0.01\textwidth}%

 \caption{Performance of different edited models as the number of edits increases across various benchmarks.}

  \label{eval_bench2}
\end{figure*}

\clearpage

\subsection{Templates}\label{Templates}

\begin{table}[H]
\setlength{\abovecaptionskip}{2pt}
\small
\centering
\label{GE_template}

\begin{tabular}{@{}l@{}} 
\toprule
\begin{minipage}{0.95\linewidth}
\begin{lstlisting}[basicstyle=\ttfamily\small,
                   breaklines=true,
                   breakautoindent=false,
                   breakindent=0pt]
Question: {question}
Answer: {answer}
Please write three different short passages (about 60 words each) 
that provide background knowledge supporting this Question-Answer pair. 

Each passage should help justify why the answer is correct. 
Separate the passages with "---".

Write them as if you were recalling from memory. 
Do not invent sources or citations.

Do NOT include any introduction or summary text before [passage_1].
Reply format:
[passage_1]
<~60 words>
---
[passage_2]
<~60 words>
---
[passage_3]
<~60 words>
\end{lstlisting}
\end{minipage}\\
\bottomrule
\end{tabular}
\caption{Template of Golden and Counter Evidence}

\end{table}

\begin{table}[H]
\setlength{\abovecaptionskip}{2pt}   
\small
\centering
\label{PE_template}
\begin{tabular}{@{}l@{}}
\toprule
\begin{minipage}{0.95\linewidth}
\begin{lstlisting}[basicstyle=\ttfamily\small,
                   breaklines=true,
                   breakautoindent=false,
                   breakindent=0pt]
[System]
You recall factual background from memory. 
Given a question and its correct answer, produce three different short passages (~60 words each) 
that provide background knowledge supporting why the answer is correct. 
Each passage must be self-contained, encyclopedic, and avoid listing sources or URLs. 
Label each as [passage_1..3] and separate with '---'.

[User]
Question: {question}
Answer: {answer}

Format EXACTLY as:
[passage_1]
<~60 words>
---
[passage_2]
<~60 words>
---
[passage_3]
<~60 words>
\end{lstlisting}
\end{minipage}\\
\bottomrule
\end{tabular}
\caption{Template of Parametric Evidence}

\end{table}

\begin{table}[H]
\setlength{\abovecaptionskip}{2pt}   
\small
\centering
\label{check}
\begin{tabular}{@{}l@{}}
\toprule
\begin{minipage}{0.95\linewidth}
\begin{lstlisting}[basicstyle=\ttfamily\small,
                   breaklines=true,
                   breakautoindent=false,
                   breakindent=0pt]
[System]
According to the given information and your knowledge, answer the question.
Information:
{INFORMATION}
Question:
{QUESTION}
Answer:
\end{lstlisting}
\end{minipage}\\
\bottomrule
\end{tabular}
\caption{Template of Consistency Check}

\end{table}

\begin{table}[H]
\setlength{\abovecaptionskip}{2pt}   
\small
\centering
\label{IE_template}
\begin{tabular}{@{}l@{}}
\toprule
\begin{minipage}{0.95\linewidth}
\begin{lstlisting}[basicstyle=\ttfamily\small,
                   breaklines=true,
                   breakautoindent=false,
                   breakindent=0pt]
[System]
You generate unrelated encyclopedic passage. 
Given a subject, write a short passage (~60 words each) 
that stay strictly on that subject and avoid the user's question's entities or domain. 
Label each as [passage].

[User]
Subject: {subject}
Forbidden keywords (do not include): {taboo_keywords}

Format EXACTLY as:
[passage]
<~60 words>
\end{lstlisting}
\end{minipage}\\
\bottomrule
\end{tabular}
\caption{Template of Irrelevant Evidence}

\end{table}

\begin{table}[H]
\setlength{\abovecaptionskip}{2pt}   
\small
\centering
\caption{Template of SA-MCQ w/ Uncertain option}
\label{MCQ_ABC_template}
\begin{tabular}{@{}l@{}}
\toprule
\begin{minipage}{0.95\linewidth}
\begin{lstlisting}[basicstyle=\ttfamily\small,
                   breaklines=true,
                   breakautoindent=false,
                   breakindent=0pt]
System:
Based on the given information and your own knowledge, please select the option that best answers the question.

Given information:
{{EVIDENCE or (none)}}

You may choose C if you are truly uncertain; otherwise choose between A or B.

User:
Question: {{QUESTION}}
A. {{PARAMETRIC_ANSWER}}
B. {{GOLD_ANSWER}}
C. I am uncertain / not sure

Answer with only the letter (A, B, or C).
\end{lstlisting}
\end{minipage}\\
\bottomrule
\end{tabular}
\end{table}

\begin{table}[H]
\setlength{\abovecaptionskip}{2pt}   
\small
\centering
\label{MCQ_ABtemplate}
\begin{tabular}{@{}l@{}}
\toprule
\begin{minipage}{0.95\linewidth}
\begin{lstlisting}[basicstyle=\ttfamily\small,
                   breaklines=true,
                   breakautoindent=false,
                   breakindent=0pt]
System:
Based on the given information and your own knowledge, please select the option that best answers the question.

Given information:
{{EVIDENCE or (none)}}

User:
Question: {{QUESTION}}
A. {{PARAMETRIC_ANSWER}}
B. {{GOLD_ANSWER}}

Answer with only the letter (A or B).
\end{lstlisting}
\end{minipage}\\
\bottomrule
\end{tabular}
\caption{Template of SA-MCQ w/o Uncertain option}

\end{table}

\begin{table}[H]
\setlength{\abovecaptionskip}{2pt}   
\small
\centering
\begin{tabular}{@{}l@{}}
\toprule
\begin{minipage}{0.95\linewidth}
\begin{lstlisting}[basicstyle=\ttfamily\small,
                   breaklines=true,
                   breakautoindent=false,
                   breakindent=0pt]
  Your job is to look at a question, a gold target, and a predicted answer, and then assign a grade of either ["CORRECT", "INCORRECT"].

The following are examples of CORRECT predicted answers.
```
Question: What are the names of Barack Obama's children?
Gold target: Malia Obama and Sasha Obama
Predicted answer 1: sasha and malia obama
Predicted answer 2: Malia and Sasha Obama are the names of Barack Obama's children.
```
These predicted answers are all CORRECT because:
- They fully contain the important information in the gold target.
- They do not contain any information that contradicts the gold target.

The following are examples of INCORRECT predicted answers.
```
Question: What are the names of Barack Obama's children?
Gold target: Malia and Sasha
Predicted answer 1: Malia.
Predicted answer 2: Malia, Sasha, and Susan.
Predicted answer 3: Malia and Sasha, Malia and Sasha, Malia and Sasha, Malia and Sasha (repeated answer)
```
These predicted answers are all INCORRECT because:
- A factual statement in the answer contradicts the gold target or contain repeated answer.

Here is a sample. Simply reply with either CORRECT or INCORRECT.

Question:\{question\} \\
Gold target: \{target\} \\
Predicted answer: \{predicted answer\} \\

According to the gold target, please grade the predicted answer of this question as one of:
A: CORRECT
B: INCORRECT

Just return the letters "A" or "B", with no text around it.  
\end{lstlisting}
\end{minipage}\\
\bottomrule
\end{tabular}
\caption{Template of LLM-as-judge}
\label{LLM-as-judge}

\end{table}

\end{document}